\documentclass{article} 
\usepackage{nips15submit_e,times}
\usepackage{url}
\usepackage{amsmath}
\usepackage{amssymb}
\usepackage{amsthm}
\usepackage{algpseudocode}
\usepackage{algorithm}
\usepackage{graphicx}
\usepackage{caption}
\usepackage{subcaption}

\newtheorem{graphconv}{Definition}

\title{Deep Convolutional Networks on Graph-Structured Data}

\author{
Mikael Henaff \\
Courant Institute of Mathematical Sciences\\
New York University\\
\texttt{mbh305@nyu.edu} \\
\And
Joan Bruna \\
University of California, Berkeley \\
\texttt{joan.bruna@berkeley.edu} \\
\AND
Yann LeCun \\
Courant Institute of Mathematical Sciences \\
New York University \\
\texttt{yann@cs.nyu.edu} \\
}

\nipsfinalcopy 

\begin{document}

\maketitle

\begin{abstract}
Deep Learning's recent successes have mostly relied on Convolutional Networks, which exploit fundamental statistical properties of images, sounds and video data: the local stationarity and multi-scale compositional structure, that allows expressing long range interactions in terms of shorter, localized interactions. However, there exist other important examples, such as text documents or bioinformatic data, that may lack some or all of these strong statistical regularities. 

In this paper we consider the general question of how to construct deep architectures with small learning complexity on general non-Euclidean domains, which are typically unknown and need to be estimated from the data. In particular, we develop an extension of Spectral Networks which incorporates a Graph Estimation procedure, that we test on large-scale classification problems, matching or improving over Dropout Networks with far less parameters to estimate.
\end{abstract}

\section{Introduction}


In recent times, Deep Learning models have proven extremely successful on a wide variety of tasks, from computer vision and acoustic modeling to natural language processing \cite{natureyann}. At the core of their success lies an important assumption on the statistical properties of the data, namely the \emph{stationarity} and the \emph{compositionality} through \emph{local} statistics, which are present in natural images, video, and speech.
 These properties are exploited efficiently by ConvNets ~\cite{krizhevsky2012,hinton12}, which are designed to extract local features that are shared across the signal domain. Thanks to this, they are able to greatly reduce the number of parameters in the network with respect to generic deep architectures, without sacrificing the capacity to extract informative statistics from the data. Similarly, Recurrent Neural Nets (RNNs) trained on temporal data implicitly assume a stationary distribution.

%
%
%
One can think of such data examples as being signals defined on a low-dimensional grid. In this case stationarity is well defined via the natural translation 
operator on the grid, locality is defined via the metric of the grid, and compositionality is obtained from downsampling, or equivalently thanks to the multi-resolution property of the grid.
However, there exist many examples of data that lack the underlying low-dimensional grid structure. 
For example, text documents represented as bags of words can be thought of as signals defined on a graph whose nodes are vocabulary terms and whose weights represent some similarity measure between terms, such as co-occurence statistics. In medicine, a patient's gene expression data can be viewed as a signal defined on the graph imposed by the regulatory network. In fact, computer vision and audio, which are the main focus of research efforts in deep learning, only represent a special case of data defined on an extremely simple low-dimensional graph. Complex graphs arising in other domains might be of higher dimension, and the statistical properties of data defined on such graphs might not satisfy the stationarity, locality and compositionality assumptions previously described.
For such type of data of dimension $N$, deep learning strategies are reduced to learning with fully-connected layers, which have $O(N^2)$ parameters, and regularization is carried out via weight decay and dropout \cite{srivastava2014dropout}.

When the graph structure of the input is known, \cite{spectralnet2013} introduced a model to generalize ConvNets using low learning complexity similar to that of a ConvNet, and which was demonstrated on simple low-dimensional graphs. In this work, we are interested in generalizing ConvNets to high-dimensional, general datasets,  and, most importantly, to the setting where the graph structure is not known a priori. In this context, learning the graph structure amounts to estimating the similarity matrix, which has complexity $O(N^2)$. One may therefore wonder whether the graph estimation followed by graph convolutions offers advantages with respect to learning directly from the data with fully connected layers. We attempt to answer this question experimentally and to establish baselines for future work. 
 
We explore these approaches in two areas of application for which it has not been possible to apply convolutional networks before: text categorization and bioinformatics. Our results show that our method is capable of matching or outperforming large, fully-connected networks trained with dropout using fewer parameters. 
Our main contributions can be summarized as follows:
\begin{itemize}
\item We extend the ideas from \cite{spectralnet2013} to large-scale classification problems, specifically Imagenet Object Recognition, text categorization and bioinformatics.
\item We consider the most general setting where no prior information on the graph structure is available, and propose unsupervised and new supervised graph estimation strategies in combination with the supervised graph convolutions.
\end{itemize}

The rest of the paper is structured as follows. Section \ref{relatedworksect} reviews similar works in the literature. Section \ref{spectralsect} discusses generalizations of convolutions on graphs, and Section \ref{graphestimsect} addresses the question of graph estimation. Finally, Section \ref{experimentssect} shows numerical experiments on large scale object recogniton, text categorization and bioinformatics.

\section{Related Work}
\label{relatedworksect}

There have been several works which have explored architectures using the so-called local receptive fields \cite{karol, coates2011selecting, ngiam2010tiled}, mostly with applications to image recognition. In particular, \cite{coates2011selecting} proposes a scheme to learn how to group together features based upon a measure of similarity that is obtained in an unsupervised fashion. However, it does not attempt to exploit any weight-sharing strategy. 

Recently, \cite{spectralnet2013} proposed a generalization of convolutions to graphs via the Graph Laplacian. By identifying a linear, translation-invariant operator in the grid (the Laplacian operator), with its counterpart in a general graph (the Graph Laplacian), one can view convolutions as the family of linear transforms commuting with the Laplacian. By combining this commutation property with a rule to find localized filters, the model requires only $O(1)$ parameters per ``feature map". However, this construction requires prior knowledge of the graph structure, and was shown only on simple, low-dimensional graphs. More recently, \cite{DBLP:journals/corr/MasciBBV15} introduced Shapenet, another generalization of convolutions on non-Euclidean domains based on geodesic polar coordinates, which was successfully applied to shape analysis, and allows comparison across different manifolds. However, it also requires prior knowledge of the manifolds. 

The graph or similarity estimation aspects have also been extensively studied in the past. For instance, \cite{ravikumar2010high} studies the estimation of the graph from a statistical point of view, through the identification of a certain graphical model using $\ell_1$-penalized logistic regression. Also, \cite{chen2014unsupervised} considers the problem of learning a deep architecture through a series of Haar contractions, which are learnt using an unsupervised pairing criteria over the features.

\section{Generalizing Convolutions to Graphs }
\label{spectralsect}
\subsection{Spectral Networks}

Our work builds upon ~\cite{spectralnet2013} which introduced spectral networks. We recall the definition here and its main properties.
A spectral network generalizes a convolutional network through the Graph Fourier Transform, which is in turn defined via a generalization of the Laplacian operator on the grid to the graph Laplacian. An input vector $x \in \mathbb{R}^N$ is seen as a a signal defined on a graph $G$ with $N$ nodes. 
\begin{graphconv}
 Let $W$ be a $N \times N$ similarity matrix representing an undirected graph $G$, and let $L = I - D^{-1/2}WD^{-1/2}$ be its graph Laplacian with $D=W\cdot {\bf 1}$ eigenvectors $U=(u_1,\dots,u_N)$. Then a \textit{graph convolution} of input signals $x$ with filters $g$ on $G$ is defined by $x \ast_G g = U^T \left( Ux \odot Ug \right)$, where $\odot$ represents a point-wise product. 
\end{graphconv}

Here, the unitary matrix $U$ plays the role of the Fourier Transform in $\mathbb{R}^d$. 
There are several ways of computing the graph Laplacian $L$ \cite{belkin2001laplacian}. In this paper, we choose the normalized version $L = I - D^{-1/2}WD^{-1/2}$, where $D$ is a diagonal matrix with entries $D_{ii} = \sum_j W_{ij}$. Note that in the case where $W$ represents the lattice, from the definition of $L$ we recover the discrete Laplacian operator $\Delta$. Also note that the Laplacian commutes with the translation operator, which is diagonalized in the Fourier basis. 
It follows that the eigenvectors of $\Delta$ are given by the Discrete Fourier Transform (DFT) matrix. 
We then recover a classical convolution operator by noting that convolutions are by definition linear operators that diagonalize in the Fourier domain (also known as the Convolution Theorem \cite{mallat1999wavelet}).

Learning filters on a graph thus amounts to learning spectral multipliers $w_g = (w_1, \dots,w_N)$ 
$$x \ast_G g := U^T ( \mbox{diag}(w_g) U x)~.$$
Extending the convolution to inputs $x$ with multiple input channels is straightforward. If $x$ is a signal with $M$ input channels and $N$ locations, we apply the transformation $U$ on each channel, and then use multipliers $w_g =  (w_{i,j}\, ;\, i \leq N~, j \leq M)$. 

However, for each feature map $g$ we need convolutional kernels are typically restricted to have small spatial support, independent of the number of input pixels $N$, which enables the model to learn a number of parameters independent of $N$. In order to recover a similar learning complexity in the spectral domain, it is thus necessary to restrict the class of spectral multipliers to those corresponding to localized filters. 

For that purpose, we seek to express spatial localization of filters in terms of their spectral multipliers. In the grid, smoothness in the frequency domain corresponds to the spatial decay, since
$$\left| \frac{\partial^k \hat{x}(\xi)}{\partial \xi^k} \right| \leq C \int |u|^k |x(u)| du~,$$
where $\hat{x}(\xi)$ is the Fourier transform of $x$.
In \cite{spectralnet2013} it was suggested to use the same principle in a general graph, by considering a smoothing kernel $\mathcal{K} \in \mathbb{R}^{N \times N_0}$, such as splines, and searching for spectral multipliers of the form
$$w_g = \mathcal{K} \tilde{w}_g~.$$

The algorithm which implements the graph convolution is described in Algorithm \ref{pseudoPSO}.

\begin{algorithm}
\caption{Train Graph Convolution Layer}
\label{pseudoPSO}
\begin{algorithmic}[1]
\State Given GFT matrix $U$, interpolation kernel $\mathcal{K}$, weights $w$. 
\State \textbf{Forward Pass:}
  \State Fetch input batch $x$ and gradients w.r.t outputs $\nabla y$.
  \State Compute interpolated weights: $w_{f'f} = \mathcal{K} \tilde{w_{f'f}}$.
  \State Compute output: $y_{sf'} = U^T\left(\sum_{f} Ux_{sf} \odot w_{f'f} \right)$.
  \State \textbf{Backward Pass:}
  \State Compute gradient w.r.t input: $\nabla x_{sf} = U^T\left(\sum_{f'} \nabla y_{sf'} \odot w_{f'f} \right)$
  \State Compute gradient w.r.t interpolated weights: $\nabla w_{f'f} = U^T\left(\sum_s \nabla y_{sf'} \odot x_{sf} \right)$
  \State Compute gradient w.r.t weights $\nabla \tilde{w_{f'f}} = \mathcal{K}^T \nabla w_{f'f}$.
\end{algorithmic}
\end{algorithm}

\subsection{Pooling with Hierarchical Graph Clustering}

In image and speech applications, and in order to reduce the complexity of the model, it is often useful to trade off spatial resolution for feature resolution as the representation becomes deeper. For that purpose, pooling layers compute statistics in local neighborhoods, such as the average amplitude, energy or maximum activation.

The same layers can be defined in a graph by providing the equivalent notion of neighborhood. In this work, we construct such neighborhoods at different scales using multi-resolution spectral clustering \cite{von2007tutorial}, and consider both average and max-pooling as in standard convolutional network architectures.

\section{ Graph Construction }
\label{graphestimsect}

Whereas some recognition tasks in non-Euclidean domains, such as those considered in \cite{spectralnet2013} or \cite{DBLP:journals/corr/MasciBBV15}, might have a prior knowledge of the graph structure of the input data, many other real-world applications do not have such knowledge. It is thus necessary to estimate a similarity matrix $W$ from the data before constructing the spectral network. In this paper we consider two possible graph constructions, one unsupervised by measuring joint feature statistics, and another one supervised using an initial network as a proxy for the estimation.

\subsection{ Unsupervised Graph Estimation }

Given data $X \in \mathbb{R}^{L \times N}$, where $L$ is the number of samples and $N$ the number of features,
the simplest approach to estimating a graph structure from the data is to consider a distance between features $i$ and $j$ given by
$$d(i,j) = \| X_i - X_j \|^2~,$$
where $X_i$ is the $i$-th column of $X$. 
While correlations are typically sufficient to reveal the intrinsic geometrical structure of images \cite{roux2008learning}, 
 the effects of higher-order statistics might be non-negligible in other contexts, especially in presence of sparsity. 
Indeed, in many situations the pairwise Euclidean distances might suffer from unnormalized measurements. Several strategies and variants 
exist to gain some robustness, for instance replacing the Euclidean distance by the $Z$-score (thus renormalizing each feature by its standard
deviation), the ``square-correlation" (computing the correlation of squares of previously whitened features), or the mutual information.

This distance is then used to build a Gaussian diffusion Kernel \cite{belkin2001laplacian} 
\begin{equation}
\label{unsupervisedkernel}
\omega(i,j) = \exp^{-\frac{d(i,j)}{\sigma^2}}~.
\end{equation}
In our experiments, we also consider the variant of self-tuning diffusion kernel \cite{zelnik2004self}
$$\omega(i,j) = \exp^{-\frac{d(i,j)}{\sigma_i \sigma_j}}~,$$
where $\sigma_i$ is computed as the distance $d(i,i_{k})$ corresponding to the $k$-th nearest neighbor $i_{k}$ of feature $i$. 
This defines a kernel whose variance is locally adapted around each feature point, as opposed to (\ref{unsupervisedkernel}) where
the variance is shared. 


The main advantage of (\ref{unsupervisedkernel}) is that it does not require labeled data. Therefore, it is possible to estimate 
the similarity using several datasets that share the same features, for example in text classification. 


\subsection{ Supervised Graph Estimation }

As discussed in the previous section, the notion of feature similarity is not well defined, as it depends on our choice of 
kernel and criteria. Therefore, in the context of supervised learning, the relevant statistics from the input signals might not correspond to 
our imposed similarity criteria. It may thus be interesting to ask for the feature similarity that best suits a particular classification task. 

A particularly simple approach is to use a fully-connected network to determine the feature similarity. Given a training set with normalized \footnote{In our experiments we simply normalized each feature by its standard deviation, but one could also whiten completely the data.} features $X \in \mathbb{R}^{L \times N}$ 
and labels $y \in \{1,\dots,C\}^L$, we initially train a fully connected network $\phi$ with $K$ layers of weights $W_1, \dots, W_K$, using standard ReLU activations and dropout. We then extract the first layer features $W_1 \in \mathbb{R}^{N \times M_1}$, where $M_1$ is the number of first-layer hidden features, and consider the distance
\begin{equation}
\label{supervisedkernel}
d_{sup}(i, j) = \| W_{1,i} - W_{1,j} \|^2~,
\end{equation}
that is then fed into the Gaussian kernel as in (\ref{unsupervisedkernel}). The interpretation is that the supervised criterion will extract through $W_1$ a collection of linear measurements that best serve the classification task. Thus two features are similar if the network decides to use them similarly within these linear measurements.  


This constructions can be seen as ``distilling" the information learnt by a first network into a kernel. In the general case where no assumptions are made on the dimension of the graph, it amounts to extracting $N^2/2$ parameters from the first learning stage (which typically involves a much larger number of parameters). If, moreover, we assume a low-dimensional graph structure of dimension $m$, then $m N$ parameters are extracted by projecting the resulting kernel into its leading $m$ directions.

Finally, observe that one could simply replace the eigen-basis $U$ obtained by diagonalizing the graph Laplacian by an arbitrary unitary matrix, which is then optimized by back-propagation together with the rest of the parameters of the model. We do not report results on this strategy, although we point out that it has the same learning complexity as the Fully Connected network (requiring $O(K N^2)$ parameters, where $K$ is the number of layers and $N$ is the input dimension).
%
%
%
%
%
%
%
%
%
%
%
%
%

\section{Experiments}
\label{experimentssect}
In order to measure the performance of spectral networks on real-world data and to explore the effect of the graph estimation procedure, we conducted experiments on three datasets from text categorization, computational biology and computer vision. All experiments were done using the Torch machine learning environment with a custom CUDA backend.

We based the spectral network architecture on that of a classical convolutional network, namely by interleaving graph convolution, ReLU and graph pooling layers, and ending with one or more fully connected layers. As noted above, training a spectral network requires an $O(N^2)$ matrix multiplication for each input and output feature map to perform the Graph Fourier Transform, compared to the efficient $O(N \text{log} N)$ Fast Fourier Transform used in classical ConvNets. We found that training the spectral networks with large numbers of feature maps to be very time-consuming and therefore chose to experiment mostly with architectures with fewer feature maps and smaller pool sizes. We found that performing pooling at the beginning of the network was especially important to reduce the dimensionality in the graph domain and mitigate the cost of the expensive Graph Fourier Transform operation.

In this section we adopt the following notation to descibe network architectures: GC$k$ denotes a graph convolution layer with $k$ feature maps, P$k$ denotes a graph pooling layer with stride $k$ and pool size $2k$, and FC$k$ denotes a fully connected layer with $k$ hidden units. In our results we also denote the number of free parameters in the network by $P_\text{net}$ and the number of free parameters when estimating the graph by $P_\text{graph}$.
\subsection{Reuters}
We used the Reuters dataset described in ~\cite{JMLR:v15:srivastava14a}, which consists of training and test sets each containing 201,369 documents from 50 mutually exclusive classes. Each document is represented as a log-normalized bag of words for 2000 common non-stop words. As a baseline we used the fully-connected network of ~\cite{JMLR:v15:srivastava14a} with two hidden layers consisting of 2000 and 1000 hidden units regularized with dropout.  

 We chose hyperparameters by performing initial experiments on a validation set consisting of one-tenth of the training data. Specifically, we set the number of subsampled weights to $k=60$, learning rate to 0.01 and used max pooling rather than average pooling. We also found that using AdaGrad ~\cite{adagrad} made training faster. All architectures were then trained using the same hyperparameters.
Since the experiments were computationally expensive, we did not train all models until full convergence. This enabled us to explore more model architectures and obtain a clearer understanding of the effects of graph construction.  

\begin{table}[H]
\caption{Results for Reuters dataset. Accuracy is shown at epochs 200 and 1500.}
\begin{center}
\begin{tabular}{|c|c|c|c|c|c|}
\hline
Graph & Architecture & $P_\text{net}$ & $P_\text{graph}$ & Acc. (200) & Acc. (1500)\\
\hline
- &FC2000-FC1000 & $6 \cdot 10^6$ & 0 &70.18 \footnotemark & 70.18 \\
Supervised & GC4-P4-FC1000 & $2\cdot 10^6$ & $2 \cdot 10^6$ & 69.41 & 70.03 \\
Supervised & GC8-P8-FC1000 & $2 \cdot 10^6$ & $2 \cdot 10^6$ & 69.15 & - \\
Supervised low rank & GC4-P4-FC1000 & $2\cdot 10^6$ & $5 \cdot 10^5$ & 69.25 & - \\
Supervised low rank & GC8-P8-FC1000 & $2 \cdot 10^6$ & $5 \cdot 10^5$ & 68.35 & - \\
Supervised & GC16-P4-GC16-P4-FC1000& $2 \cdot 10^6$ & $2 \cdot 10^6$ & 69.04 & - \\
Supervised &GC64-P8-GC64-P8-FC1000 & $2 \cdot 10^6$ & $2 \cdot 10^6$ & 69.09 & - \\
RBF kernel & GC4-P4-FC1000 & $2\cdot 10^6$ & $2 \cdot 10^6$ & 67.85 & - \\
RBF kernel & GC8-P8-FC1000 & $2 \cdot 10^6$ & $2 \cdot 10^6$ & 66.95 & - \\
RBF kernel & GC16-P4-GC16-P4-FC1000 & $2 \cdot 10^6$ & $2 \cdot 10^6$ & 67.16 & - \\
RBF kernel & GC64-P8-GC64-P8-FC1000 & $2 \cdot 10^6$ & $2 \cdot 10^6$ & 67.42 & - \\
RBF kernel (local)& GC4-P4-FC1000 & $2\cdot 10^6$ & $2 \cdot 10^6$ & 68.56 & - \\
RBF kernel (local) & GC8-P8-FC1000 & $2 \cdot 10^6$ & $2 \cdot 10^6$ & 67.66 & - \\
\hline
\end{tabular}
\end{center}
\label{reuterstable}
\end{table}
\footnotetext{this is the maximum value before the fully connected starts overfitting}

Note that our architectures are designed so that they factor the first hidden layer of the fully connected network across feature maps and a subsampled graph, trading off resolution in the graph domain for resolution across feature maps. The number of inputs into the last fully connected layer is always the same as for the fully-connected network. The idea is to reduce the number of parameters in the first layer of the network while avoiding too much compression in the second layer. 
We note that as we increase the tradeoff between resolution in the graph domain and across features, there reaches a point where performance begins to suffer. This is especially pronounced for the unsupervised graph estimation strategies. When using the supervised method, the network is much more robust to the factorization of the first layer. Table \ref{reuterstable} compares the test accuracy of the fully connected network and the GC4-P4-FC1000 network. Figure \ref{merckfigure}-left shows that the factorization of the lower layer has a beneficial regularizing effect. 

\begin{figure}
       \centering
        \includegraphics[width=0.25\textwidth]{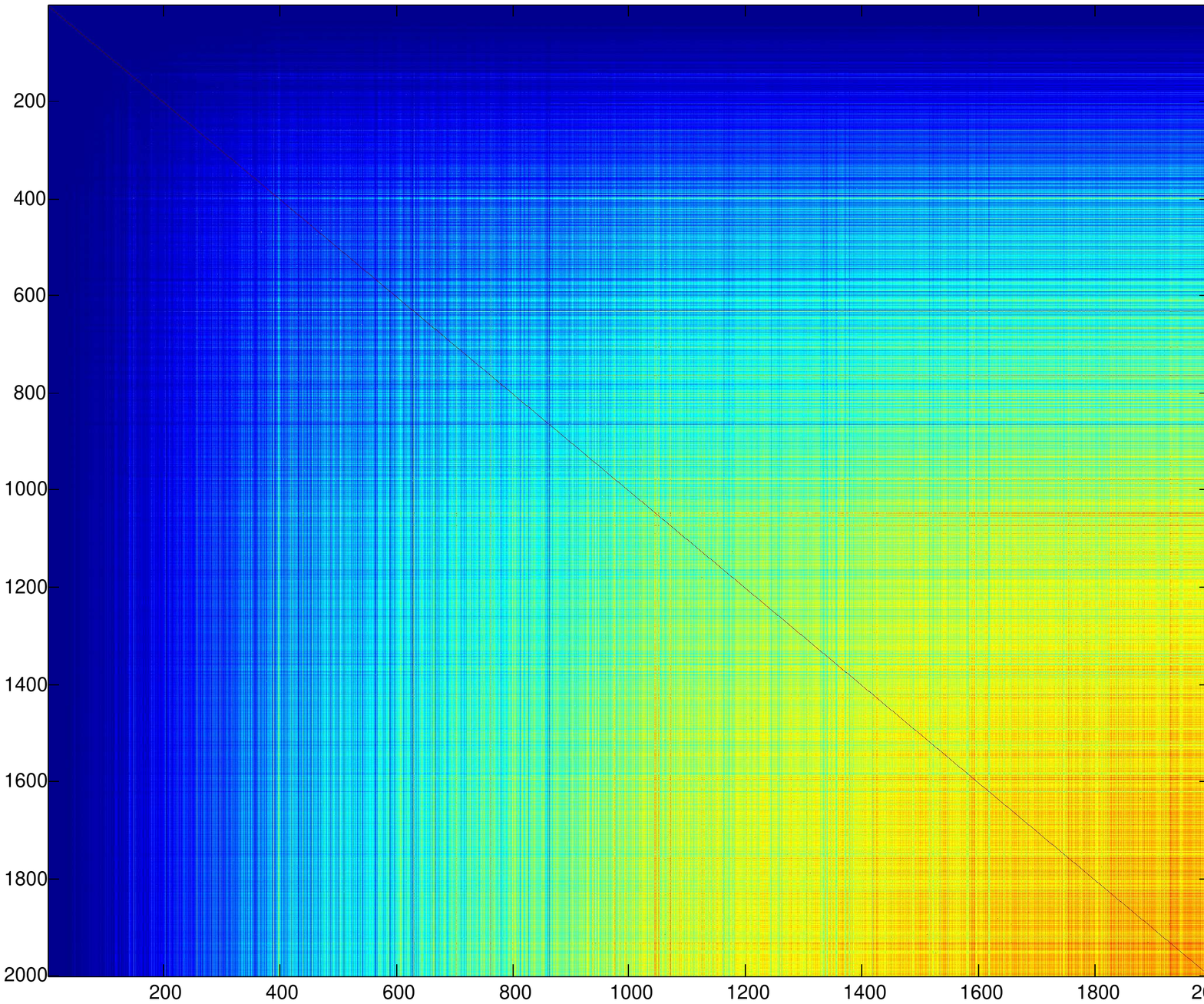}
         \includegraphics[width=0.25\textwidth]{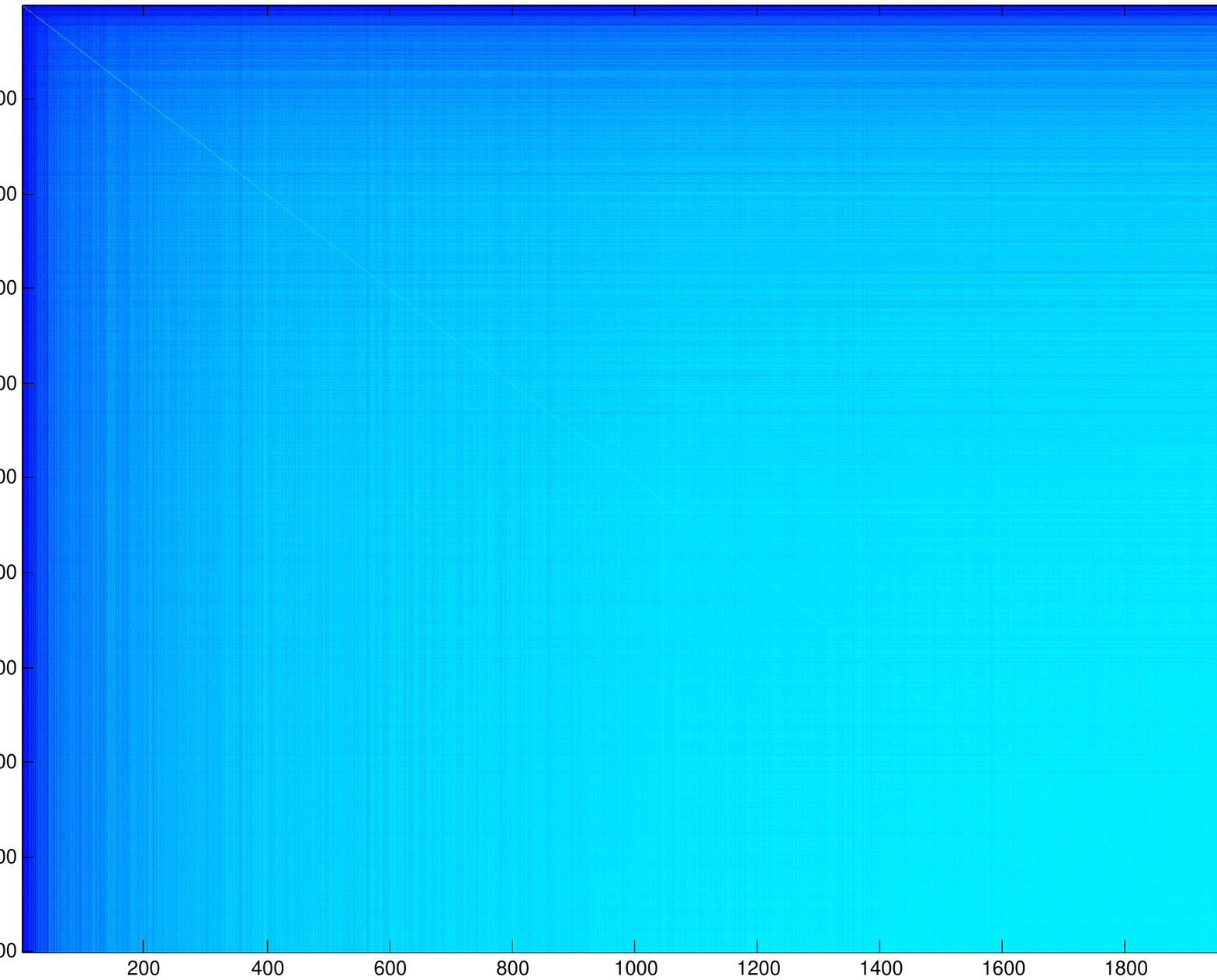} \\
          \includegraphics[width=0.25\textwidth]{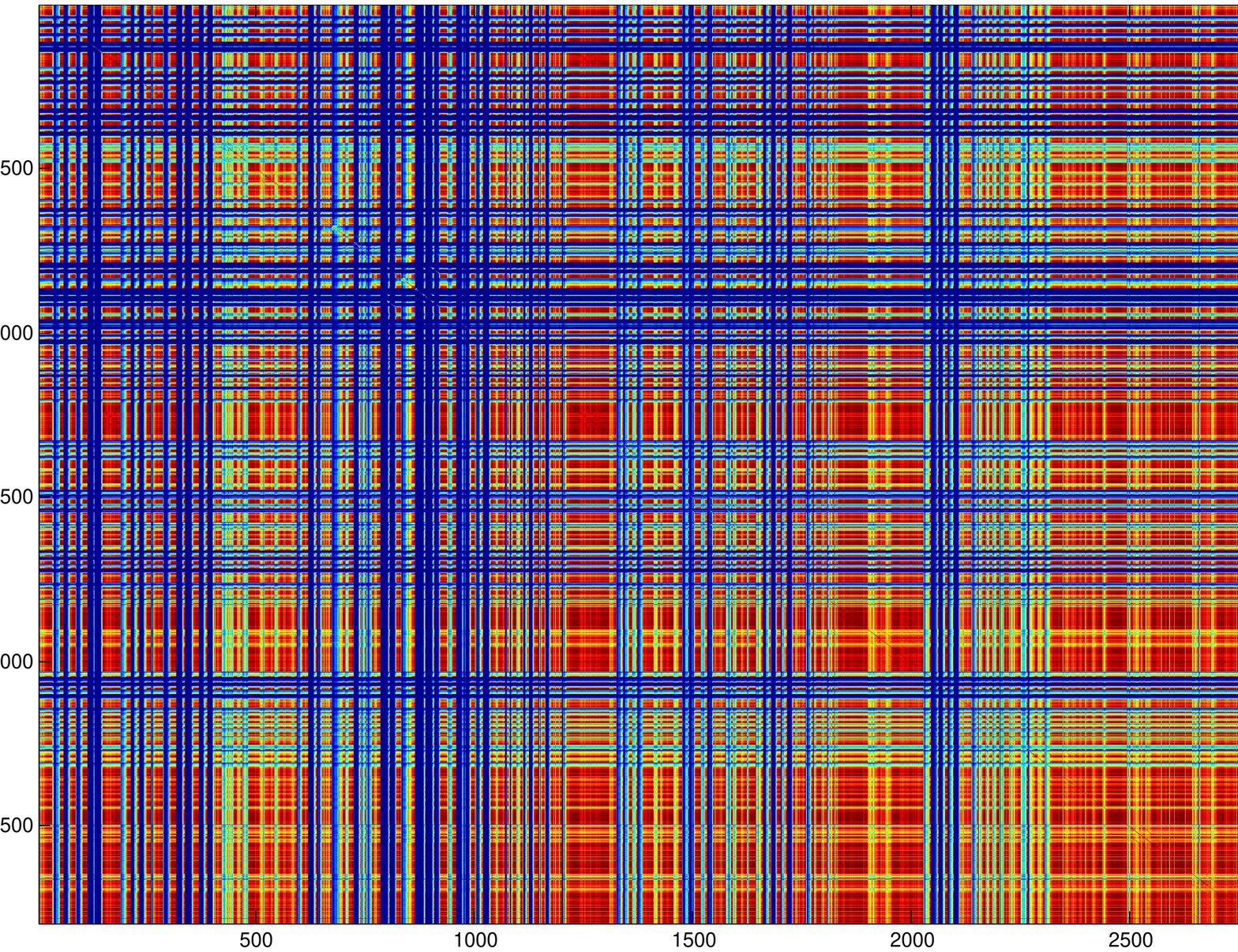}
          \includegraphics[width=0.25\textwidth]{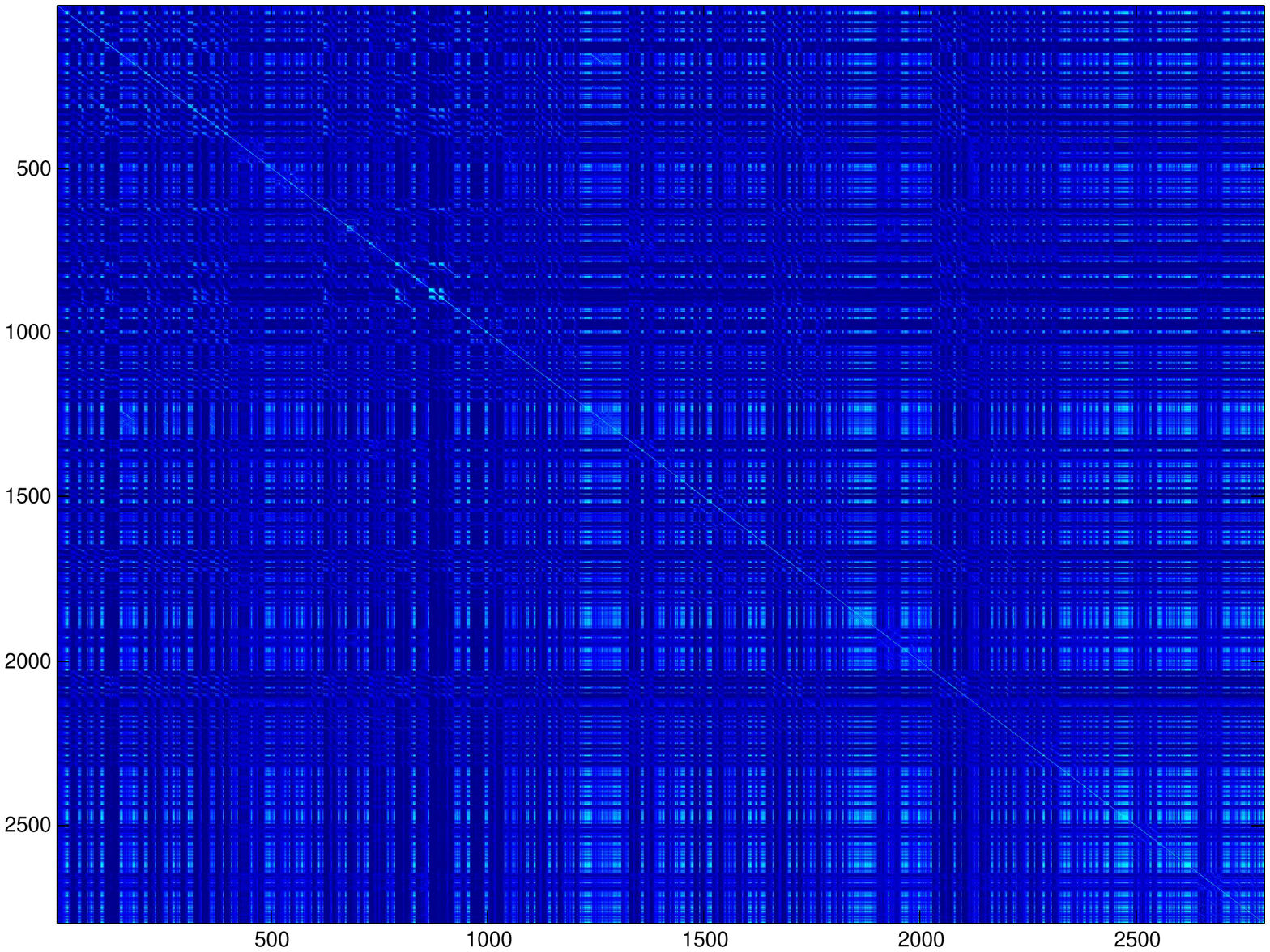}
       \caption{Similarity graphs for the Reuters (top) and Merck DPP4 (bottom) datasets. Left plots correspond to global $\sigma$, right plots to local $\sigma$.}\label{fig:animals}
\end{figure}

%

\subsection{Merck Molecular Activity Challenge}

The Merck Molecular Activity Challenge is a computational biology benchmark where the task is to predict activity levels for various molecules based on the distances in bonds between different atoms. For our experiments we used the DPP4 dataset which has 8193 samples and 2796 features. We chose this dataset because it was one of the more challenging and was of relatively low dimensionality which made the spectral networks tractable. As a baseline architecture, we used the network of ~\cite{Ma:JournalChem2015} which has 4 hidden layers and is regularized using dropout and weight decay. We used the same hyperparameter settings and data normalization recommended in the paper.   

As before, we used one-tenth of the training set to tune hyperparameters of the network. For this task we found that $k=40$ subsampled weights worked best, and that average pooling performed better than max pooling. Since the task is to predict a continuous variable, all networks were trained by minimizing the Root Mean-Squared Error loss. Following ~\cite{Ma:JournalChem2015}, we measured performance by computing the squared correlation between predictions and targets.

\begin{figure}[h]
\begin{center}
\includegraphics[width=0.4\textwidth]{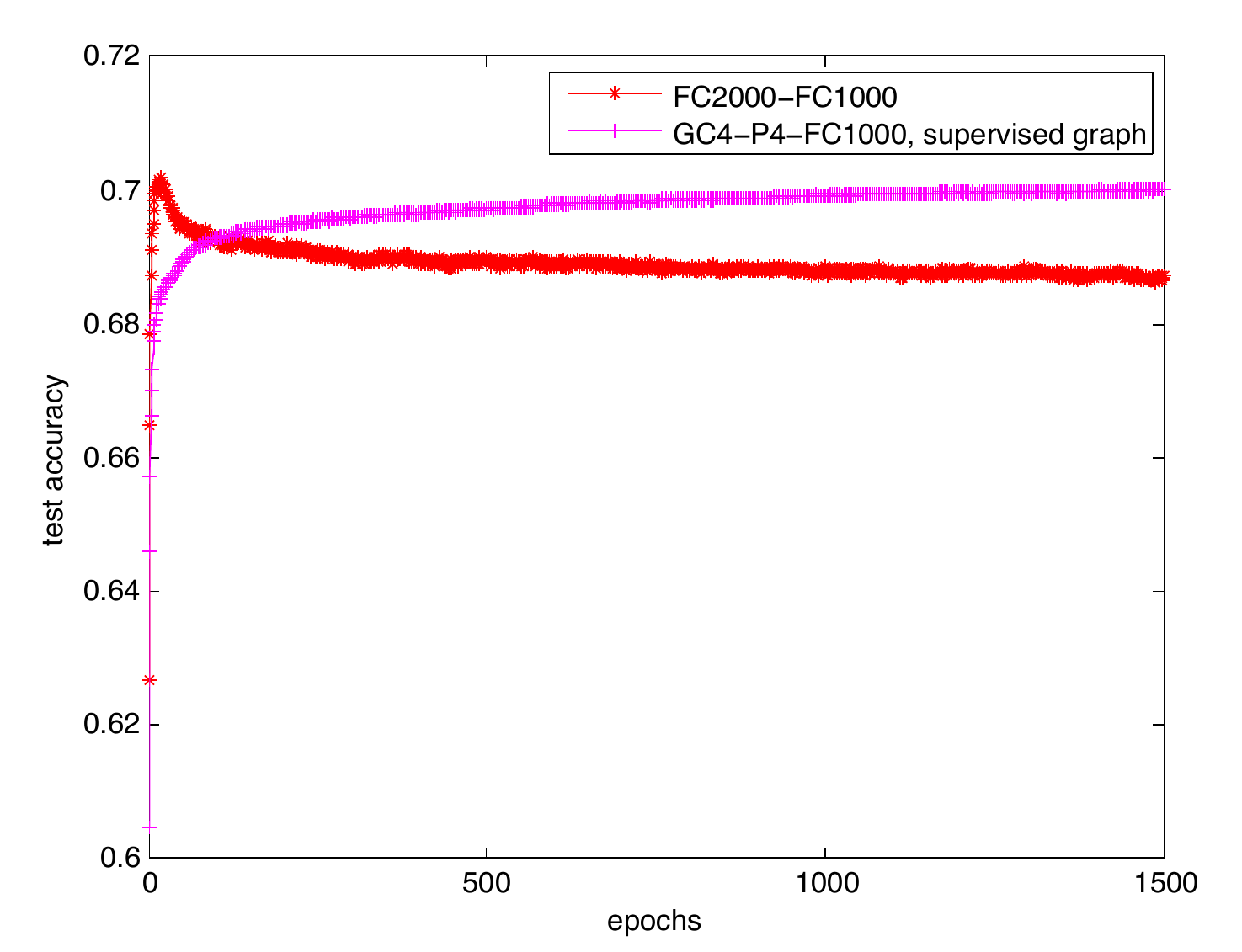}~
\includegraphics[width=0.4\textwidth]{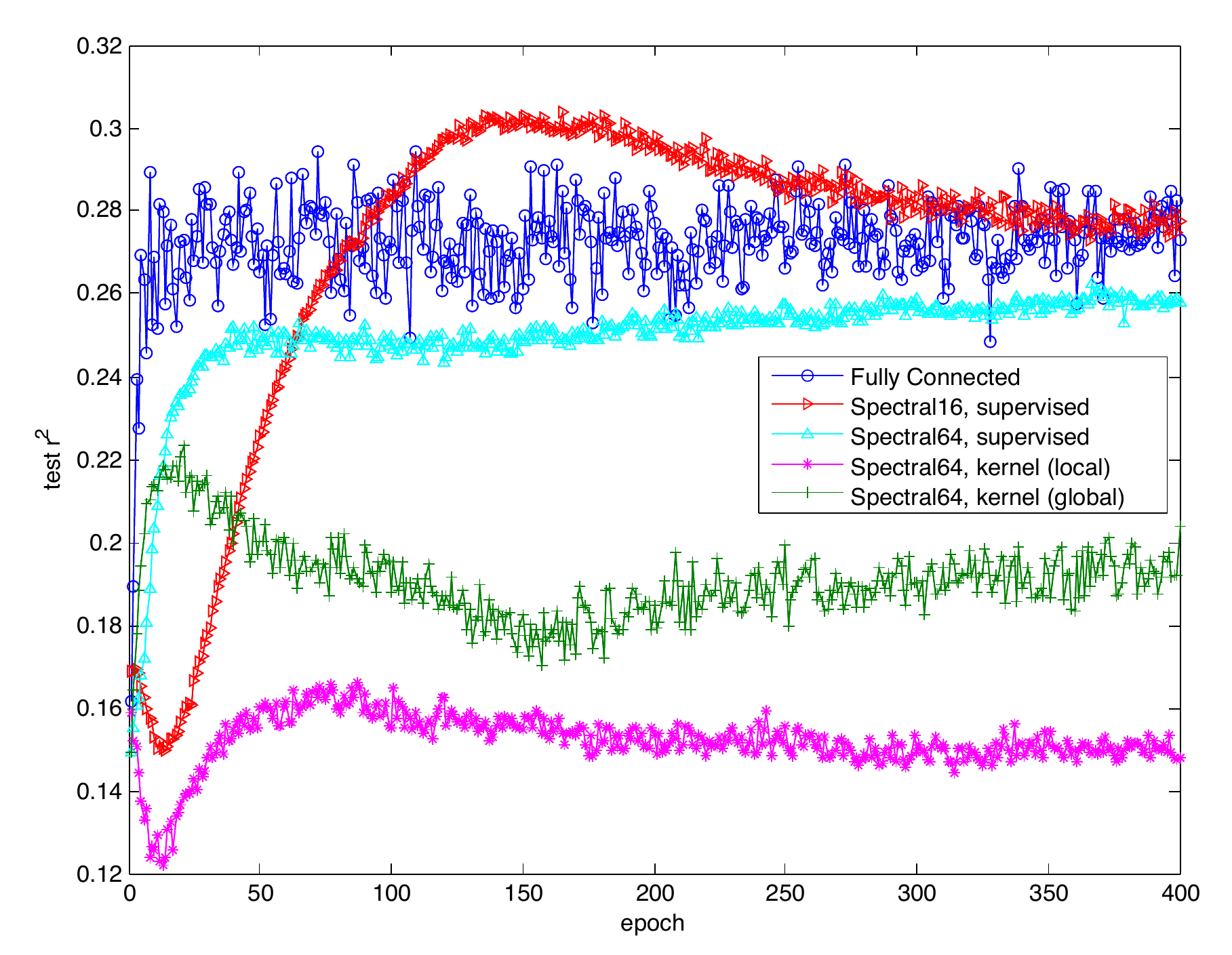}
\caption{Evolution of Test accuracy. Left: Reuters dataset, Right: Merck dataset.}
\end{center}
\end{figure}
\label{merckfigure}

\begin{table}[H]
\caption{Results for Merck DPP4 dataset.}
\begin{center}
\begin{tabular}{|c|c|c|c|c|}
\hline
Graph & Architecture & $P_\text{net}$ & $P_\text{graph}$ & $R^2$\\ 
\hline
- &FC4000-FC2000-FC1000-FC1000 & $22.1 \cdot 10^6$ & 0 & 0.2729 \\
Supervised & GC16-P4-GC16-P4-FC1000-FC1000 & $3.8\cdot 10^6$ & $3.9 \cdot 10^6$ & 0.2773 \\
Supervised & GC64-P8-GC64-P8-FC1000-FC1000 & $3.8\cdot 10^6$ & $3.9 \cdot 10^6$ &0.2580 \\
RBF Kernel & GC64-P8-GC64-P8-FC1000-FC1000 & $3.8\cdot 10^6$ & $3.9 \cdot 10^6$ &0.2037 \\
RBF Kernel (local) & GC64-P8-GC64-P8-FC1000-FC1000 & $3.8\cdot 10^6$ & $3.9 \cdot 10^6$  &0.1479 \\
\hline
\end{tabular}
\end{center}
\end{table}

We again designed our architectures to factor the first two hidden layers of the fully-connected network across feature maps and a subsampled graph, and left the second two layers unchanged. As before, we see that the unsupervised graph estimation strategies yield a significant drop in performance whereas the supervised strategy enables our network to perform similarly to the fully-connected network with much fewer parameters. This indicates that it is able to factor the lower-level representations in such a way as to retain useful information for the classification task.

Figure \ref{merckfigure}-right shows the test performance as the models are being trained. We note that the Merck datasets have test set samples assayed at a different time than the samples in the training set, and thus the distribution of features is typically different between the training and test sets. Therefore the test performance can be a significantly noisy function of the train performance. However, the effect of the different graph estimation procedures is still clear. 

\subsection{ImageNet}

In the experiments above our graph construction relied on estimation from the data. 
To measure the influence of the graph construction compared to the filter learning in the graph frequency domain, we performed the same experiments on the ImageNet dataset for which the graph is already known, namely it is the 2-D grid. The spectral network was thus a convolutional network whose weights were defined in the frequency domain using frequency smoothing rather than imposing compactly supported filters.
 Training was performed exactly as in Figure 1, except that the linear transformation was a Fast Fourier Transform. 

Our network consisted of 4 convolution/ReLU/max pooling layers with 48, 128, 256 and 256 feature maps, followed by 3 fully-connected layers each with 4096 hidden units regularized with dropout. We trained two versions of the network: one classical convolutional network and one as a spectral network where the weights were defined in the frequency domain only and were interpolated using a spline kernel. Both networks were trained for 40 epochs over the ImageNet dataset where input images were scaled down to $128 \times 128$ to accelerate training.  

\begin{table}[H]
\caption{ImageNet results}
\begin{center}
\begin{tabular}{|c|c|c|c|}
\hline
Graph & Architecture & Test Accuracy (Top 5) & Test Accuracy (Top 1)\\
\hline
2-D Grid & Convolutional Network & 71.854 & 46.24\\
2-D Grid & Spectral Network & 71.998 & 46.71\\
\hline
\end{tabular}
\end{center}
\end{table}

\begin{figure}[h]
\centering
 \includegraphics[width=0.5\textwidth]{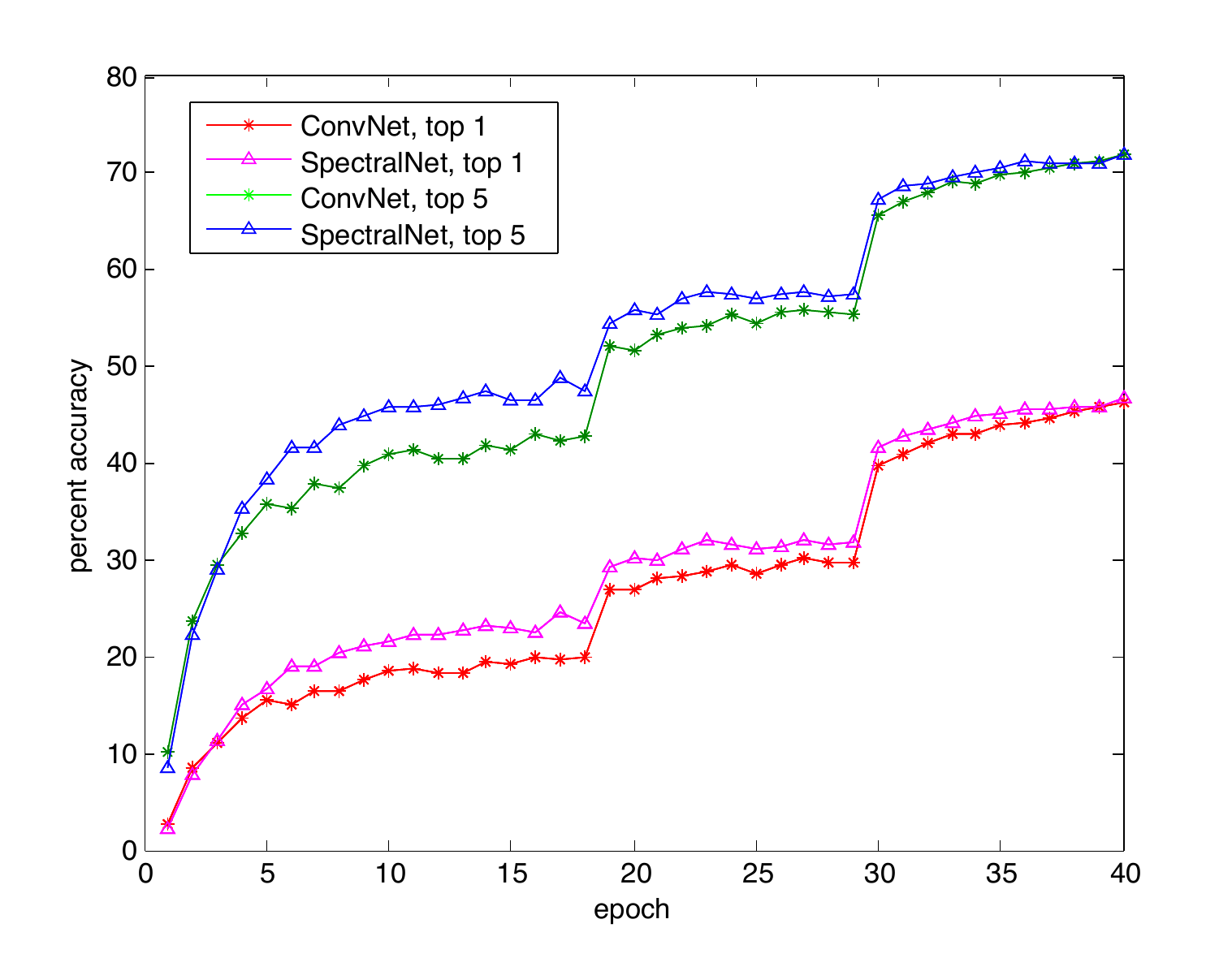}
 \caption{ConvNet vs. SpectralNet on ImageNet.}
\end{figure}

We see that both models yield nearly identical performance. Interstingly, the spectral network learns faster than the ConvNet during the first part of training, although both networks converge around the same time. This requires further investigation.

\section{Discussion}

ConvNet architectures base their appeal and success on their 
ability to produce highly informative local statistics using low learning complexity
and avoiding expensive matrix multiplications. This motivated us to 
consider generalizations on high-dimensional, unstructured data.

When the statistical properties of the input satisfy both stationarity 
and composotionality, spectral networks have a learning complexity 
of the same order as Convnets. In the general setting where no
prior knowledge of the input graph structure is known, our model 
requires estimating the similarities, a $O(N^2)$ operation, but making
the model deeper does not increase learning complexity as much as 
the general Fully Connected architectures. Moreover, in contexts
where feature similarities can be estimated using unlabeled data (such as
word representations), our model has less parameters to learn from labeled data.

However, as our results demonstrate, 
their extension poses significant challenges:
\begin{itemize}
\item Although the learning complexity requires $O(1)$ parameters per feature map, 
the evaluation, both forward and backward, requires a multiplication by the Graph Fourier Transform, 
which costs $O(N^2)$ operations. This is a major difference with respect to traditional ConvNets, which 
require only $O(N)$. Fourier implementations of Convnets \cite{mathieu2013fast, DBLP:journals/corr/VasilacheJMCPL14} bring the complexity to $O(N \log N)$ 
thanks again to the specific symmetries of the grid. An open question is whether one can find approximate eigenbasis
of general Graph Laplacians using Givens' decompositions similar to those of the FFT. 
\item Our experiments show that when the input graph structure is not known a priori, graph estimation is
the statistical bottleneck of the model, requiring $O(N^2)$ for general graphs and $O(M N)$ for $M$-dimensional graphs.
Supervised graph estimation performs significantly better than unsupervised graph estimation based on low-order moments. Furthermore, we have verified that
the architecture is quite sensitive to graph estimation errors. In the supervised setting, this step can be viewed in terms of
a Bootstrapping mechanism, where an initially unconstrained network is self-adjusted to become more localized and with weight-sharing. 
\item Finally, the statistical assumptions of stationarity and compositionality are not always verified. In those situations, the constraints
imposed by the model risk to reduce its capacity for no reason. One possibility for addressing this issue is to insert Fully connected 
layers between the input and the spectral layers, such that data can be transformed into the appropriate statistical model. Another strategy, that is
left for future work, is to relax the notion of weight sharing by introducing instead a commutation error $\| W_i L - L W_i \|$ with the graph Laplacian, 
which puts a soft penalty on transformations that do not commute with the Laplacian, instead of imposing exact commutation as is the case in the spectral net. 
\end{itemize}


\bibliography{references}{}
\bibliographystyle{plain}

\end{document}